\documentclass[11pt,a4paper]{article}
\usepackage[hyperref]{emnlp2020}
\usepackage{times}
\usepackage{latexsym}

\usepackage{amsmath}
\usepackage{amssymb}
\usepackage{graphicx}

\usepackage{arydshln}
\usepackage{booktabs}

\usepackage{url}

\newcommand\blfootnote[1]{%
  \begingroup
  \renewcommand\thefootnote{}\footnote{#1}%
  \addtocounter{footnote}{-1}%
  \endgroup
}

\usepackage{microtype}

\aclfinalcopy 
\def\aclpaperid{859} 

\title{Transition-based Parsing with Stack-Transformers}

\author{Ram\'{o}n Fernandez Astudillo$^\dagger$ ~~~ Miguel Ballesteros$^*$$^\diamondsuit$ \\ \textbf{Tahira Naseem}$^\dagger$ ~~~ 
\textbf{Austin Blodget}$^{*\heartsuit}$ ~~~ \textbf{Radu Florian}$^\dagger$\\
$^\dagger$IBM Research ~~~ $^\diamondsuit$Amazon AI ~~~ 
$^\heartsuit$Georgetown University\\
\texttt{\footnotesize{\{ramon.astudillo,tnassem,raduf\}@ibm.com}}}

\date{}

\begin{document}
\maketitle
\begin{abstract}
Modeling the parser state is key to good performance in transition-based parsing. Recurrent Neural Networks considerably improved the performance of transition-based systems by modelling the global state, e.g. stack-LSTM parsers, or local state modeling of contextualized features, e.g. Bi-LSTM parsers. Given the success of Transformer architectures in recent parsing systems, this work explores modifications of the sequence-to-sequence Transformer architecture to model either global or local parser states in transition-based parsing. We show that modifications of the cross attention mechanism of the Transformer considerably strengthen performance both on dependency and Abstract Meaning Representation (AMR) parsing tasks, particularly for smaller models or limited training data.\blfootnote{$^*$Miguel's and Austin's contributions were carried out while at IBM Research.}
\end{abstract}

\section{Introduction}
\label{section:intro}

Transition-based Parsing transforms the task of predicting a graph from a sentence into predicting an action sequence of a state machine that produces the graph \cite{nivre2003efficient,nivre2004incrementality,Kubler:2009:DP:1538443,Henderson:2013:MJP:2576217.2576223}. These parsers are attractive for their linear inference time and interpretability, however, their performance hinges on effective modeling of the parser state at every decision step. 

Parser states typically comprise two memories, a buffer and a stack, from which tokens can be pushed or popped \cite{Kubler:2009:DP:1538443}. Traditionally, parser states were modeled using hand selected \textit{local} features pertaining only to the words on the top of the stack or buffer \cite[inter-alia]{nivre2007maltparser,zhang-nivre-2011-transition}. With the widespread use of neural networks, \textit{global} models of the parser state such as the stack-LSTM~\cite{dyer2015transition} allowed encoding the entire buffer and stack. It was later shown that \textit{local} features of the stack and buffer extracted from contextual word representations, such as Bi-LSTMs, could outperform global modeling \cite{kiperwasser-goldberg-2016-simple,dozat2016deep}.

With the rise of the Transformer model \cite{vaswani2017attention}, various approaches have been proposed that leverage this architecture for parsing \cite{kondratyuk201975,kulmizev2019deep,mrini2019rethinking,ahmad-etal-2019-difficulties,cai2020amr}. In this work we revisit the local versus global paradigms of state modeling in the context of sequence-to-sequence Transformers applied to action prediction for transition-based parsing. Similarly to previous works for RNN sequence to sequence \cite{liu-zhang-2017-encoder,zhang-etal-2017-stack}, we propose a modification of the cross-attention mechanism of the Transformer to provide global parser state modeling. We analyze the role of local versus global parser state modeling, stack and buffer modeling, effects model size as well as task complexity and amount of training data.

Results show that local and global state modeling of the parser state yield more than $2$ percentage points absolute improvement over a strong Transformer baseline, both for dependency and Abstract Meaning Representation (AMR) parsing. Gains are also particularly large for smaller train sets and smaller model sizes, indicating that parser state modeling, can compensate for both. Finally, we improve the AMR transition-based oracle \cite{ballesteros2017amr}, yielding best results for a transition-based system and second overall.

\section{Global versus Local Parser State}
\label{section:state}
\begin{figure*}
\centering
  \includegraphics[width=16cm]{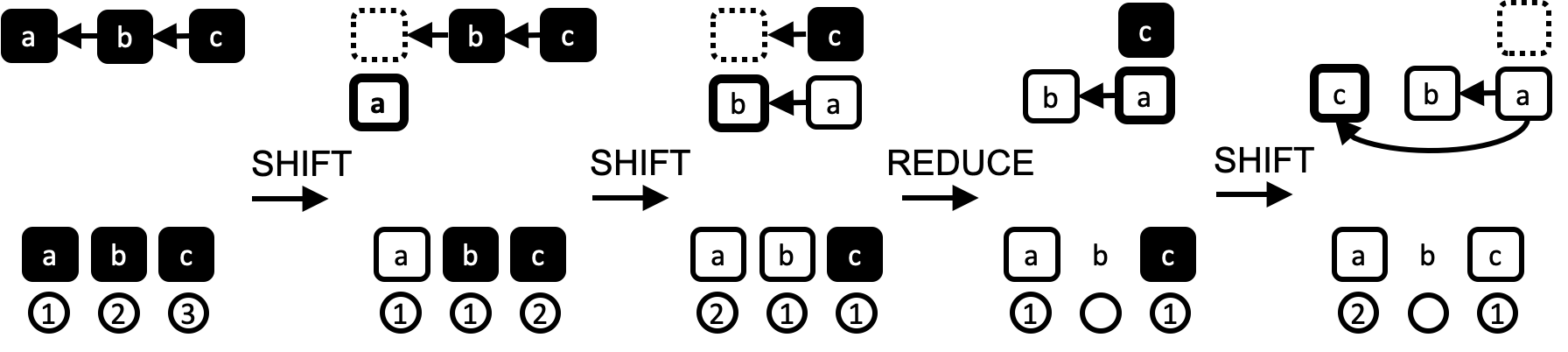}
  \caption{Encoding of buffer and stack for action sequence $a = \{\mbox{SHIFT}, \mbox{SHIFT}, \mbox{REDUCE}, \mbox{SHIFT}\}$ and sentence $w = \{a, b, c\}$. The stack-LSTM is at the top, with hidden states representation of buffer (black) and stack (white) displayed. The stack-Transformer is at the bottom, with masks for cross-attention heads attending buffer (black) and stack (white) displayed. Circles indicate extra cross-attention positions relative to stack and buffer.}
  \label{fig:stack-encodings}
\end{figure*}

Given pair of sentence $\mathbf{w} = w_1, w_2 \cdots w_N$ and graph $\mathbf{g}$, transition-based parsers learn an action sequence $\mathbf{a} = a_1, a_2 \cdots a_T$, that applied to a state machine yields the graph $\mathbf{g} = M(\mathbf{a}, \mathbf{w})$. Actions of the state machine generally move words from a buffer, that initially contains the entire sentence, to a stack. Components of the graph, such as edges or nodes, are created by applying transformations to words in the stack. The correct action sequence is given by an oracle $\mathbf{a} = O(\mathbf{w}, \mathbf{g})$, which is generally rule-based. In principle, one could learn the sentence to action mapping $\mathbf{w} \rightarrow \mathbf{a}$ as a sequence to sequence problem
\begin{equation}
p(\mathbf{a} \mid \mathbf{w}) = \prod_{t=1}^Tp(a_t \mid \mathbf{a}_{<t}, \mathbf{w})\nonumber,
\end{equation}
similarly to e.g. Machine Translation. In practice, this approach does not accurately represent the parser state and thus shows limited performance. The parser state at step $t$ is defined implicitly by $(\mathbf{a}_{<t}, \mathbf{w})$. This translates to an explicit state at step $t$ where the stack contains some tokens about to be processed, sometimes along with new composed vector representations, and the buffer contains the remainder of tokens in the sentence. Buffer and stack increase (push) or decrease (pop) their size dynamically with each time step as shown in Fig.~\ref{fig:stack-encodings}.

The transition-based formalism relies heavily on the explicit representation of the state i.e. buffer and stack configurations. Prior to widespread use of Neural Networks, \textit{local} features limited to top of the stack and buffer already achieved good performances \cite[inter-alia]{nivre2007maltparser,zhang-nivre-2011-transition}. The introduction of stack-LSTMs \cite{dyer2015transition} made possible modeling the \textit{global} state of the parser by separately encoding action history $\mathbf{a}_{<t}$, and the dynamically changing stack and buffer with LSTMs \cite{hochreiter1997long}. In addition to this, stack-LSTMs used the transition-based formalism to recursively build vector representations of sub-graphs, similarly to a graph neural network. 

Another well known LSTM model is the Bi-LSTM feature parser \cite{kiperwasser-goldberg-2016-simple,dozat2016deep}. In this case, a contextual representation of the sentence is first built with a Bi-LSTM $\mathbf{h}=\mathrm{BiLSTM}(\mathbf{w})$. At each time step $t$ the stack configuration determined by $\mathbf{a}_{<t}$ is used to select the elements from $\mathbf{h}$ corresponding to words on the top of the stack and buffer. Although the features utilize local information of the buffer and stack, the use of a strong contextual representation proved to be sufficient and this remains one of the most widely used forms of parsing today.

\section{Modeling Parser State in Transformers}
\label{section:state}

\subsection{From stack-LSTMs to stack-Transformers}

In transition-based parsers, at a given time step $t$, input tokens $\mathbf{w}$ may be on the buffer, stack or reduced. As displayed in Fig.~\ref{fig:stack-encodings} (top), to encode this state stack-LSTMs unroll LSTMs over the stack and buffer following their respective word order, which can be different from the sentence's token order. If an element is added to the buffer or stack, it is only necessary to unroll one additional LSTM cell. If an element is removed under a pop operation (e.g. \textrm{REDUCE}), stack-LSTMs move back a pointer to reuse previously computed hidden states. This allows efficient encoding of the dynamically changing stack and buffer.

Unlike LSTMs, Transformers \cite{vaswani2017attention} encode sequences through an attention mechanism \cite{bahdanau-etal-2015-neural} as a weighted sum of tokens plus position embeddings. One can take advantage of this mechanism to replace LSTMs with Transformers for stack and buffer encoding. Since Transformers just sum token representations, under a pop operation elements can be masked out and there is no need for a pointer. Furthermore, since Transformers use multiple heads one can have separate modeling of stack and buffer by specializing two heads of the attention mechanism, see Fig.~\ref{fig:stack-encodings} (bottom), while the other heads remain free. 

In practical terms, we modify the cross-attention mechanism of the Transformer decoder. For example, for the head attending the stack, the score function between action history encoding $b_t$ (query) and hidden representation of word $h_i$ (key) is given by \vspace{-.2cm}
\begin{equation}
    e_{ti}^{\mbox{\footnotesize stack}} = \frac{b_t W^Q \left((h_i + p_{ti}^{\mbox{\footnotesize stack}}) W^K \right)^T}{\sqrt{d}} + m_{ti}^{\mbox{\footnotesize stack}}\nonumber,
\end{equation}
where $m_{ti}$ is a $\{-\infty, 0\}$ mask, $p_{ti}$ are the position embeddings for elements in the stack, $\mathbf{h} = f(\mathbf{w})$ is the output of the Transformer encoder. The attention would be computed from the score function as 
\begin{equation}
\alpha_{ti} = \mathrm{softmax}(\mathbf{e}_{t})_i\nonumber
\end{equation}
Both mask and positions change for each word and time-step as the parser state changes, but they imply little computation overhead and can be precomputed for training. Henceforth this modification will be referred to as stack-Transformer.

\subsection{Labeled SHIFT Multi-task}

It is common practice for transition-based systems to add an additional Part of Speech (POS) or word prediction task \cite{bohnet2012transition}. This is achieved by labeling the \textrm{SHIFT} action, that moves a word from the buffer to the stack, with the word's tag. This decorated actions become part of the action history $\mathbf{a}_{<t}$, which was expected to give better visibility into stack/buffer content and exploit Transformer's attentional encoding of history. In initial experiments, POS tags produced a small improvement while word prediction led to performance decrease. It was observed, however, that prediction of only $100-300$ most frequent words, leaving \textrm{SHIFT} undecorated otherwise, led to large performance increases. This is thus the method reported in the experimental setup as alternative parser state modeling.

\section{Experiments and Results}
\label{section:expsetup}

\begin{table*}[ht!]
\centering
\scalebox{0.89}{
\begin{tabular}{l|c|c|c|c}
\hline
&\multicolumn{2}{c|}{Penn-Treebank}&AMR 1.0&AMR 2.0\\
\hline
Model & UAS & LAS & {Smatch}  & {Smatch}\\ 
\hline
a) vanilla sequence-to-sequence Transformer              & 93.9\footnotesize{$\pm$0.2} & 92.0\footnotesize{$\pm$0.3} & 70.5 \footnotesize{$\pm$0.1} & 77.7\footnotesize{$\pm$0.1}\\
b) label top $100$ SHIFT multi-task                      & 95.2\footnotesize{$\pm$0.1} & 93.5\footnotesize{$\pm$0.1} & 74.9\footnotesize{$\pm$0.3}  & 79.0\footnotesize{$\pm$0.1}\\
c) 1 head attends stack, 1 head attends buffer           & {\bf 95.7\footnotesize{$\pm$0.1}}& {\bf 94.1\footnotesize{$\pm$0.2}} & {\bf 76.3\footnotesize{$\pm$0.0}}  & {\bf 79.5\footnotesize{$\pm$0.2}}\\
\hline
d) 1 head attends stack, 1 head attends buff. + stack/buff. positions  & 94.8\footnotesize{$\pm$0.1} & 92.7\footnotesize{$\pm$0.2} & 70.7\footnotesize{$\pm$0.2} & 70.4\footnotesize{$\pm$2.6}\\
e) 1 head attends entire buffer                          & {\bf 95.8\footnotesize{$\pm$0.1}}& {\bf 94.0\footnotesize{$\pm$0.1}} & 76.2\footnotesize{$\pm$0.1} & {\bf 79.7\footnotesize{$\pm$0.0}}\\
f) 1 head attends entire stack                           & 95.5\footnotesize{$\pm$0.1}  & 93.8\footnotesize{$\pm$0.1} & 75.9\footnotesize{$\pm$0.2} & 79.4\footnotesize{$\pm$0.2}\\
g) 1 head attends top two words of the buffer            & 95.7\footnotesize{$\pm$0.1}  & 93.8\footnotesize{$\pm$0.1} & 75.9\footnotesize{$\pm$0.2} & 79.4\footnotesize{$\pm$0.1}\\
h) 1 head attends top two words of the stack             & 95.4\footnotesize{$\pm$0.1}  & 93.8\footnotesize{$\pm$0.1} & 76.1\footnotesize{$\pm$0.2} & 79.4\footnotesize{$\pm$0.2}\\
\hspace{3mm} + label top $100$ SHIFT multi-task          & 95.4\footnotesize{$\pm$0.1}  & 93.8\footnotesize{$\pm$0.1} & {\bf 76.5\footnotesize{$\pm$0.1}} & 79.4\footnotesize{$\pm$0.0}\\
\hline
small a) sequence-to-sequence Transformer                & 91.0\footnotesize{$\pm$0.6}  & 88.5\footnotesize{$\pm$0.6} & 66.5\footnotesize{$\pm$0.1}  & 74.4\footnotesize{$\pm$0.2}\\
small b) label top $100$ SHIFT multi-task                & 94.6\footnotesize{$\pm$0.2}  & 92.5\footnotesize{$\pm$0.2} & 72.6\footnotesize{$\pm$0.1}& 76.7\footnotesize{$\pm$0.1}\\
small c)  1 head attends stack, 1 head attends buffer    & {\bf 95.5\footnotesize{$\pm$0.1}}  & {\bf 93.7\footnotesize{$\pm$0.1}} & {\bf 75.8\footnotesize{$\pm$0.0}}& {\bf 79.1\footnotesize{$\pm$0.1}}\\
\hline
\end{tabular}
}
\caption{Dev-set performance for PTB (simpler parsing task), AMR2.0 (complex parsing task) and AMR1.0 (one third of AMR2.0 train data). Top: encoding parser state through multi-task or multi-head attention modification. Middle: different encodings of global/local state by multi-head attention modification. Bottom: Effect of small model size ($4$ layers). All models use fixed RoBERTa-base contextualized embeddings, checkpoint average and beam $10$. All results are average of $3$ different random seeds with standard deviation indicated with $\pm$.}
\label{table:dev}
\end{table*}

To test the proposed approach, different parsing tasks were selected. Dependency parsing in the English-Treebank, is well known and well resourced ($40$K sentences). The AMR2.0 semantic parsing task is more complex, encompassing named entity recognition, word sense disambiguation and co-reference among other sub-tasks, also well resourced ($36$K sentences). AMR1.0 has around $10$K sentences and can be considered as AMR with limited train data. 

The dependency parsing setup followed \citet{dyer2015transition}, in the setting with no POS tags. This has only \textrm{SHIFT}, \textrm{LEFT-ARC(label)}, and \textrm{RIGHT-ARC(label)} base action with a total of $82$ different actions. Results were measured in terms of (Un)labeled Attachment Scores (UAS/LAS).

The AMR setup followed \citet{ballesteros2017amr}, which introduced new actions to segment text and derive nodes or entity sub-graphs. In addition, we use the alignments and wikification from \citet{naseem2019rewarding}. Unlike previous works, we force-aligned the unaligned nodes to neighbouring words and allowed attachment to the leaf nodes of entity sub-graphs, this increased oracle Smatch from $93.7$ to $98.1$ and notably improved model performance. We therefore provide results for the \newcite{naseem2019rewarding} oracle for comparison. Both previous works predict a node creation action and then a node label, or call a lemmatizer if no label is found. Instead, we directly predicted the label and added \textrm{COPY} actions to construct node names from lemmas\footnotemark\footnotetext{We used https://spacy.io/ as lemmatizer} or surface words, resulting in a maximum of $9$K actions. Node label predictions were limited to those seen during training for the word on the top of the stack. Results were measured in Smatch \cite{cai-knight-2013-smatch} using the latest version \textrm{1.0.4}\footnotemark\footnotetext{Note that bug fixes in Smatch seem to yield $0.3$ improvements against its 2019 version.}. 

\begin{table}[!t]
\centering
\scalebox{0.89}{
\begin{tabular}{l|c|c}
\hline
\toprule
Model                                                & UAS        & LAS  \\
\midrule
\hline
\citet{dozat2016deep}                                & 95.7       & 94.0 \\ 
\citet{fernandez-gonzalez-gomez-rodriguez-2019-left} & 96.0       & 94.4\\
\citet{mohammadshahi2020recursive}$^\beta$           & 96.7       & 95.0 \\ 
\citet{mrini2019rethinking}$^X$                      & {\bf 97.3}       & {\bf 96.3} \\ 
\hline
a) Transformer                                      & 94.4\footnotesize{$\pm$0.1} & 92.6\footnotesize{$\pm$0.2}\\
b) Transformer + (mul.-task)                        & 96.0\footnotesize{$\pm$0.1} & 94.4\footnotesize{$\pm$0.1} \\
e) Stack-Transformer (buff)                         & 96.3\footnotesize{$\pm$0.0} & 94.7\footnotesize{$\pm$0.0}\\
c) Stack-Transformer                                & 96.2\footnotesize{$\pm$0.1} & 94.7\footnotesize{$\pm$0.0}\\
\hline
\end{tabular}
}
\caption{Test-set performance for Table~\ref{table:dev} selections and prior art on the English Penn-Treebank.}
\label{table:dep-test}
\end{table}

Regarding model implementation, all models were implemented on the fairseq toolkit and trained with only minor modifications over the MT model hyper-parameters \cite{ott2018scaling}. This used cross-entropy training with learning rate $5e^{-4}$, inverse square root scheduling with min. $1e^{-9}$, $4000$ warm-up updates with learning rate $1e^{-7}$, and maximum $3584$ tokens per batch. Adam parameters $0.9$ and $0.98$, label smoothing was reduced to $0.01$\footnotemark\footnotetext{see \url{https://github.com/pytorch/fairseq/tree/master/examples/scaling_nmt}}. All models used $6$ layers of encoding and decoding with size $256$ and $4$ attention heads, except the normal Transformers in AMR, which performed better on a $3$/$8$ layer configuration instead of $6$/$6$. To study the effect of model size, \textit{small} versions of all models using a $2$/$2$ configuration were also tested. 

We used RoBERTa-base \cite{liu2019roberta} embeddings without fine-tuning as input, averaging wordpieces to obtain word representations. Weight averaging of the best $3$ checkpoints \cite{junczys-dowmunt-etal-2016-amu} and beam $10$ were used in all models. This improves results at most by $0.4/0.8$ points for AMR2.0/AMR1.0 with no significant differences across models. Models were trained for a fixed number of epochs, selecting the best model on validation by either LAS or Smatch. A maximum epoch number of $80-120$ was set to guarantee a margin of $5$ epochs from best model to last epoch. No other hyper-parameters were changed across models or tasks. Training took at most $6$h on a Nvidia Tesla v100 GPU. It should be noted that this is around $10$ times faster than our Pytorch stack-LSTM implementation for the same data. The labeled SHIFT strategy used the $100$ most frequent words. 

\section{Analysis of Results}
\label{section:results}

Table~\ref{table:dev} compares the standard Transformer, with and without multi-task with the stack-Transformer, its components, and smaller versions of all models. Comparing LAS and Smatch, stack-transformer provides around $2$ points improvement against Transformer on PTB and AMR2.0, and $0.5$ points improvement against its multi-task version (a-c). This improvement becomes sensibly larger for the smaller train set AMR1.0 with $5.8$ and $1.4$ point gains over the Transformer and its multi-task version respectively. Differences are also larger for the $4$ layer version of the models. Under this setting, the stack-Transformer looses only $0.4$ points against a $12$ layer model in AMR2.0. In this same setting, the Transformer and its multi-task version loose $3.3$ and $2.3$ points respectively, pointing to the fact that modeling parser state compensates for less training data or smaller models.

Regarding ablation of the stack-Transformer components, the use of stack/buffer positions seems clearly detrimental (d) in all scenarios. This was a consistent pattern across various variants for which we do not report numbers such as sinusoidal versus learnable positions and reducing the position range to top three of the stack and buffer. One possible explanation is that positions varying after each time step may be hard to learn, particularly if injected directly in the decoder. It is also worth noting, than the combination of multi-task and stack-Transformer produced little improvement or was even detrimental pointing to their similar role. Results for the weakest of the stack-Transformer variants are provided (h).

Comparing across different attention modifications (e-h), most methods perform similarly although there seems to be some evidence for global (full buffer, full stack) variants being more performant. Modeling of the buffer seems also more important than modeling of the stack. One possible explanation for this is that, since the total number of heads is kept fixed, it may be more useful to gain an additional free head than modeling the stack content. Furthermore without recursive representation building, as in stack-LSTMs, the role of the stack can be expected to be less important. 

Tables \ref{table:dep-test} and \ref{table:amr-test} compare with prior works. Pre-trained embeddings used are indicated as XL-net-large$^X$ \cite{yang2019xlnet}, BERT base$^\beta$ and large$^B$ \cite{devlin-etal-2019-bert}, Graph Recategorization, which utilizes an external entity recognizer \cite{lyu2018amr,zhang2019broad} as (G.R.) and a$^*$ indicates the \newcite{naseem2019rewarding} oracle.

Overall, the stack-Transformer is competitive against recent works particularly for AMR, likely due to the higher complexity of the task. Compared to prior AMR systems, it is worth noting the large performance increase against stack-LSTM \cite{naseem2019rewarding}, while sharing a similar oracle and embeddings and not using reinforcement learning fine-tuning. The stack-Transformer also matches the best reported AMR system \cite{cai2020amr} on AMR1.0 without graph recategorization, but using RoBERTa instead of BERT embeddings and provided the second best reported scores on the higher resourced AMR2.0 \footnotemark\footnotetext{Code available under \url{https://github.com/IBM/transition-amr-parser/}}. 

\begin{table}[!t]
\centering
\scalebox{0.89}{
\begin{tabular}{l|c|c}
\hline
Model             & AMR1.0 & AMR2.0\\
\hline
\hline
\citet{lyu2018amr} (G.R.)                         & 73.7 & 74.4 \\  
\citet{naseem2019rewarding}$^B$      & -    & 75.5 \\ 
\citet{zhang2019broad} (G.R.) $^B$                &71.3  & 77.0 \\
\citet{cai2020amr} $^\beta$                       &74.0  & 78.7 \\
\citet{cai2020amr} (G.R.) $^\beta$                &{\bf 75.4}  & {\bf 80.2} \\
\hline
a$^*$) Transformer            &68.8\footnotesize{$\pm$0.1}&75.9\footnotesize{$\pm$0.3}\\
a) Transformer                &69.2\footnotesize{$\pm$0.2} & 77.2\footnotesize{$\pm$0.2}\\
b) Transformer (mul.-task)  &74.0\footnotesize{$\pm$0.2} & 78.0\footnotesize{$\pm$0.1} \\
e) Stack-Transformer (buff)   &75.1\footnotesize{$\pm$0.3} & 78.8\footnotesize{$\pm$0.1}\\
c) Stack-Transformer          &{\bf 75.4\footnotesize{$\pm$0.0}} & 79.0\footnotesize{$\pm$0.1}\\

\hline
\end{tabular}
}
\caption{Test-set performance for Table~\ref{table:dev} selections and prior art on the AMR1.0 and AMR2.0 in terms of Smatch.}
\label{table:amr-test}
\end{table}


\section{Related Works}
\label{section:related}

While inspired by stack-LSTMs \cite{dyer2015transition}, the stack-Transformer lacks their elegant recursive composition, where representations for partial graph components are added to the stack and used in subsequent representations. It allows, however, to model the global parser state in a simple way that is easy to parallelize, and shows large performance gains against stack-LSTMs on AMR. The proposed modified attention mechanism, could also be interpreted as a form of feature-based parser \cite{kiperwasser-goldberg-2016-simple}, where the parser state is used to select encoder representations, integrated into a Transformer sequence to sequence model.

The modification of the attention mechanism to reflect the parse state has been applied in the past to RNN sequence-to-sequence models. \newcite{liu-zhang-2017-encoder} propose the use of a boundary to separate stack and buffer attentions. While simple, this precludes the use of SWAP actions needed for AMR parsing and non-projective parsing. \newcite{zhang-etal-2017-stack} mask out reduced words and add a bias to the attention weights for words in the stack. While being the closest to the proposed technique, this method does not separately model stack and buffer nor retains free attention heads, which we consider a fundamental advantage. We also provide evidence that modeling the parser state still produces gains when using pre-trained Transformer embeddings and provide a detailed analysis of components. Finally, RNN \cite{ma-etal-2018-stack} and self-attention \cite{ahmad-etal-2019-difficulties} Stack-Pointer networks sum encoder representations based on local graph structure, which can be interpreted as masked uniform attention over $3$ words and is related to the previous methods.

\section{Conclusions}

We have explored modifications of sequence-to-sequence Transformers to encode the parser state for transition-based parsing, inspired by stack-LSTM's global modeling of the parser state. While simple, these modifications consistently provide improvements against a normal sequence to sequence Transformer in transition-based parsing, both for dependency parsing and AMR parsing tasks. Results also point to the benefits of modeling the parser state as a way to compensate for limited training resources or limitation in model sizes. 


\end{document}